x_

# Convolutional Graph Auto-encoder: A Deep Generative Neural Architecture for Probabilistic Spatio-temporal Solar Irradiance Forecasting

Mahdi Khodayar, *Student Member, IEEE*, Saeed Mohammadi, *Student Member, IEEE,* Mohammad Khodayar, *Senior Member, IEEE* and Jianhui Wang, *Senior Member, IEEE,* Guangyi Liu*, Senior Member, IEEE*

*Abstract*— Machine Learning on graph-structured data is an important and omnipresent task for a vast variety of applications including anomaly detection and dynamic network analysis. In this paper, a deep generative model is introduced to capture continuous probability densities corresponding to the nodes of an arbitrary graph. In contrast to all learning formulations in the area of discriminative pattern recognition, we propose a scalable generative optimization/algorithm theoretically proved to capture distributions at the nodes of a graph. Our model is able to generate samples from the probability densities learned at each node. This probabilistic data generation model, i.e. convolutional graph auto-encoder (CGAE), is devised based on the localized first-order approximation of spectral graph convolutions, deep learning, and the variational Bayesian inference. We apply our CGAE to a new problem, the spatio-temporal probabilistic solar irradiance prediction. Multiple solar radiation measurement sites in a wide area in northern states of the US are modeled as an undirected graph. Using our proposed model, the distribution of future irradiance given historical radiation observations is estimated for every site/node. Numerical results on the National Solar Radiation Database show state-of-the-art performance for probabilistic radiation prediction on geographically distributed irradiance data in terms of reliability, sharpness, and continuous ranked probability score.

## I. Introduction

IN recent years, the rapid exhaustion of fossil fuel sources, the environmental pollution concerns, and the aging of the developed power plants are considered as crucial global concerns. As a consequence, the renewable energy resources including wind and solar have been rapidly integrated into the existing power grids. The reliability of power systems depends on the capability of handling expected and unexpected changes and disturbances in the production and consumption, while maintaining quality and continuity of service. The variability and stochastic behavior of photovoltaic (PV) power caused by the solar radiation uncertainty leads to major challenges including voltage fluctuations, as well as local power quality and stability issues [2]. Hence, accurate solar irradiance forecasting for PV estimation is required for effective operation of power grids [3]. The studies in the area of solar irradiance and PV power forecasting are mainly categorized into three major classes:

1) The persistence model is applied as a baseline that assumes the irradiance values at future time steps is equal to the same values at the forecasting time. Due to such a strong smoothness assumption, the persistence scheme is only effective for intra-hour applications [2].

2) Physical models employ physical processes to estimate the future solar radiation values using astronomical relationships [4], meteorological parameters, and numerical weather predictions (NWPs) [5]. In [6], an hourly-averaged day-ahead PV forecasting approach is presented based on least squares optimization of NWPs using global horizontal irradiance (GHI) and the zenith angle. Some NWPs make use of the clear sky radiation modeled by earth-sun geometry [7] or panel tilt/orientation along with several meteorological parameters such as temperature or wind speed [8]. Other works apply cloud motion vector (CMV) frameworks [9] for accurate short-term predictions, using static cloud images [10], satellite images [11], or the sensor networks [12].

3) Statistical and Artificial intelligence (AI) techniques are recently presented for a number of solar irradiance and PV power estimation/regression problems. As discussed in [13], the non-stationary and highly nonlinear characteristics of solar radiation time series lead to the superiority of AI approaches over the traditional statistical models. Machine learning algorithms are employed as target function approximators, to estimate future solar irradiance or PV power. Highly nonlinear regression methodologies including ANNs [14] and support vector machines/regression (SVM/R) [15] have been employed for short-term purposes. [15] presents a benchmarking of supervised neural networks, Gaussian processes and support vector machines for GHI predictions. In [16, 17] a bootstrapping approach is presented to estimate uncertainties involved in prediction of wind/solar time series. Here, a number of Extreme Learning Machine (ELM) ANNs are trained as regression models using resampled training data. The

Mahdi Khodayar, Saeed Mohammadi, Mohammad Khodayar and Jianhui Wang are with the Department of Electrical Engineering of Southern Methodist University, Dallas, TX, USA (mahdik@smu.edu, smohammadi@smu.edu, mkhodayar@smu.edu and jianhui@smu.edu), Guangyi Liu is with Global Energy Interconnection Research Institute North America (GEIRI North America or GEIRINA), San Jose, CA, USA (email: guangyi.liu@geirina.net)

uncertainties in solar/wind data and the model uncertainties are modeled as two classes of uncertainties to provide probabilistic predictions. This model has low generalization capability as both uncertainties are associated with a strong prior knowledge that forces the uncertainties to be Gaussian. [18] employs k-nearest neighborhood (k-NN) method to find days with similar weather condition. Kernel Density Estimation (KDE) is further applied to estimate the probability density function (PDF) of PV for the neighbors of k-NN. [19] proposed a probabilistic prediction models based on linear Quantile Regression (QR), combining the point prediction obtained by a deterministic forecasting approach, with the information retrieved from ground measurements.

In this paper, a new problem, probability distribution learning in graph-structured data, is solved as a recent pattern recognition challenge. First, generative modeling (learning mathematical patterns from a dataset for the aim of generating new samples under the observed data distribution) is introduced as an optimization problem where the probability of observed data in a given dataset is maximized. Then, our novel graph learning model, Convolutional Graph Auto-encoder (CGAE), is presented that is mathematically proved to learn continuous probability density functions from the nodes in any arbitrary graph. Our CGAE is defined based on the first-order approximation of graph convolutions (for learning a compact representation from an input graph) and standard function approximation (more specifically, deep neural architectures with high generalization capacity). The proposed deep learning model is able to generate new samples corresponding to each node, after observing historical graph-structured data, while learning the nodal distributions.

In this study, the problem of spatio-temporal probabilistic solar radiation forecasting is presented as a graph distribution learning problem solved by the CGAE. First, a set of solar measurement sites in a wide area is modeled as an undirected graph, where each node represents a site and each edge reflects the correlation between historical solar data of its corresponding nodes/sites. CGAE is applied to the graph in order to learn the distributions corresponding to the solar data at each site/node. Our CGAE is mathematically guaranteed to efficiently generate samples corresponding to the future solar irradiance values. The samples generated by this model result in a probabilistic solar radiation forecast for the future time step.

The key contributions of this work are: 1) Our CGAE is the first model devised in the area of machine learning, for the problem of nodal distribution learning in graph-structured data. The presented work is a universal model/algorithm that can be applied to any arbitrary graph for the probability approximation problems. 2) This is the first study of generative modeling for the prediction of renewable resources. Previous works including all ANNs [14, 16, 17], regression [19], and kernel methods such as SVMs and SVRs [15], as well as all KNN-based methodologies [18] follow discriminative modeling, and no generative modeling was introduced in the literature. As shown by the mathematical proof, our generative model leads to accurately understanding the underlying distribution of solar data, while discriminative modeling cannot provide such capability. 3) A spatio-temporal probabilistic forecasting framework is presented that makes use of the knowledge obtained from neighboring solar sites to enhance the prediction reliability and sharpness. 4) In contrast to previous ANN-based approaches [14, 16, 17] that merely apply shallow architectures, i.e. models with small number of hidden layers, here, our model is able to have as many latent layers as it needs in order to provide the optimal generalization capability to increase the validation accuracy. As a result, the generalization capability and the learning capacity of our proposed deep network is much higher than previous works. Increasing the number of layers in previous models, even with the existence of a regularization error term, is infeasible as it would lead to the vanishing gradient problem. However, here, we solve the issue of having low gradient magnitude that arise in shallow architectures.

The paper is organized as follows: In Section II the problem of probabilistic solar irradiance forecasting is defined. In section III, first, our proposed generative modeling paradigm is defined mathematically. Then, our CGAE model is formulated and its application for solving the forecasting problem is explained. Theoretical guarantee of the proposed methodology is available in this section. Section IV explains the performance metrics and shows numerical results on a large dataset. Finally, the conclusions and future works on generative modeling are presented in Section V.

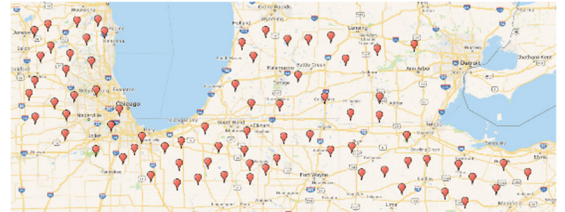
Fig. 1. Location of 75 solar sites in National Solar Radiation Database

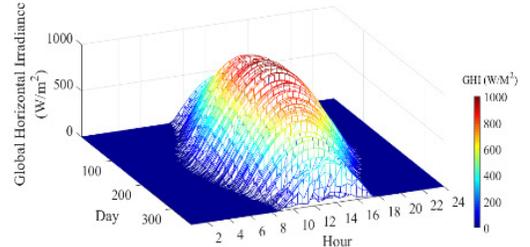
Fig. 2. Global Horizontal Irradiance measurments in 2015 for a solar site 14

## II. Problem Formulation for Probabilistic Solar Irradiance Forecasting

The solar irradiance time series measured at 75 solar sites in northern states of the US near the Lake Michigan, are collected in the National Solar Radiation Database (NSRD) [31] by the National Renewable Energy Laboratory. Fig. 1 depicts the location of the solar sites where the spatio-temporal solar radiation data is collected. The data at each site contains the GHI time series with 30-min intervals from 1998 up to 2016. Fig. 2 is the plot of GHI values at site 14 in 2015. As shown here, GHI increases from 8:00 to 13:00, and then, decreases until it reaches zero from about 18:00 to 20:00. Generally speaking, we have larger GHI around the day 200 (mid-July), and as we go further, the GHI declines.

The spatio-temporal data is modeled as an undirected graph where each node represents a solar site and each edge reflects the correlation between corresponding nodes/sites. Let us define a graph $G = (V_G, E_G)$ where $V_G$ is the set of nodes $v_i$ $i = 1, 2, ..., n$ and $E_G$ is the set of edges $e_{kl}$ connecting $v_k$ to $v_l$. Here, at each time step $t$, each node $v_i$ contains a GHI time series $T(v_i, t)$ corresponding to the historical GHI data used as the input to the forecasting model in order to predict some future GHI value $v_i^*(t' = t + k)$ with forecast horizon length $k > 0$. The problem is to learn a conditional probability distribution $P^*(V^*(t') | \pi)$ with future GHI tensor $V^*(t') = <v_1^*(t'), v_2^*(t'), ..., v_n^*(t')>$ and historical GHI tensor $\pi = <T(v_1, t), T(v_2, t), ..., T(v_n, t)>$. Considering a training set $TS$ that contains $|TS|$ historical examples $(\pi_j, V_j^*(t'))$ $1 \le j \le |TS|$, we need to estimate $P^*$ using the observed $\pi_j$ and $V_j^*(t')$ in the $j$-th training example.

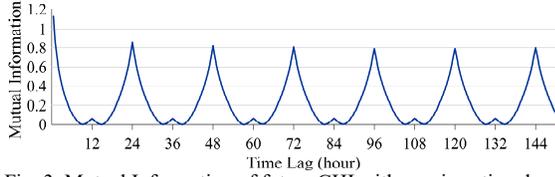

Fig. 3. Mutual Information of future GHI with previous time lags

The data of 1998-2015 is considered for training our model while the 2016 dataset is used as a test set to evaluate our method. Fig. 3. shows the mutual information between a GHI value at time $\tilde{t}$ with previous time steps $\tilde{t} - l$ with lag $1 \le l \le 300$ for the GHI time series of 1998-2015. As shown in this plot, the GHI values are more correlated with their most recent lags as well as the time lags near $l \in \{24, 48, 72, 96, 120, 144\}$. In this study, in order to make the information in $T(v_i, t)$ useful for the estimation of $P^*$, we define $T(v_i, t)$ for each node $i$ to be the GHI values corresponding to the lags where the mutual information is equal or greater than some threshold $\tau \ge 0$. Here, $\tau$ is a hyper parameter for our model.

## III. Proposed Generative Learning Formulation for Nodal Probability Density Estimation in Graphs

### A. Generative Learning for PDF approximation

Here, our problem is to capture a probability distribution $P(X)$ over $n$-dimensional data points $X$ in a potentially high dimensional vector space $\mathcal{X} \subseteq \mathbb{R}^n$. In fact, we want to be able to generate many samples $X^*$ as close as possible to $X$. As the complexity of the dependencies between variables of $\mathcal{X}$ grows, the difficulty of learning the true $P(X)$ increases. Hence, we define a "latent variable"-based model in which the hidden random vector $z \in Z$ embody the major characteristics of $P(X)$ (e.g. the PDF of the future GHI, or any desired nodal PDF in a graph-structured data). More specifically, $z$ is sampled following some unknown distribution $P(z)$ over the high dimensional space $Z$. To justify that our approach is generative (i.e. the model can generate samples $X^*$), we ensure that there exists at least one configuration $\hat{z} \in Z$ that causes the model to generate some sample $\hat{X}$ in $\mathcal{X}$. Assuming a family of deterministic functions $f(z; \theta)$ with parameters $\theta \in \Theta$, each "latent variable-parameter" pair is mapped to a sample in $\mathcal{X}$ using $f : Z \times \Theta \to \mathcal{X}$. We find an optimal $\theta^* \in \Theta$ such that when $z \sim P(z)$, the value of $X^* = f(z; \theta = \theta^*)$ is as close as possible to some $X \in \mathcal{X}$. In other words, the probability of $f$ creating an output $X^*$ similar to the observed data $X$ is maximized; hence, our optimization is written as:

$$\theta^* = \underset{\theta}{\arg\max} \left[ P(X) = \int f(z; \theta) P(z) dz \right] \quad (1)$$

Since $f$ is a random variable, $P(X)$ in (1) can be written as:

$$P(X) = \int P(X | z; \theta) P(z) dz \quad (2)$$

As shown in (1), generating $X$ depends on latent vector $z$. Using the Maximum Likelihood framework, if the model converges to the solution $\theta^*$, our generative model is likely to produce $X^*$. Here, $f(z; \theta)$ is defined as a Gaussian distribution $P(X | z; \theta) = N(X | f(z; \theta), \sigma^2 * I)$ with mean $f$ and a diagonal covariance matrix with entries computed using the hyper parameter $\sigma$ as the standard deviation.

In order to solve the optimization (1)-(2), $z$ should be mathematically defined. Moreover, an estimation for the integral in (1) should be provided. Our main goal is to learn variable $z$ automatically; that is, we opt to avoid describing the dependencies between the dimensions of $Z$, as no prior knowledge is available/required to solve the problem. Thus, the latent vector is set to $z \sim N(0, I)$ considering Theorem (1):

**Theorem (1):** *In any space $\Lambda$, any complicated probability density function over samples can be modeled using a set of $dim(\Lambda)$ random variables with normal distribution, mapped through a high capacity function.*

As a consequence, an approximator can be learned to map $z$ to some required (desired) hidden variable $\xi$ further mapped to $X \in \mathcal{X}$, to maximize the likelihood of samples $X$ in the dataset $D$. Here, our $f$ is modeled by an ANN as a standard function approximator capable of learning highly nonlinear target functions using multiple hidden layers. The first layers of these architectures provides a non-linear mapping from $z \in Z$ (with a predefined simple distribution as discussed in this section) to $\xi$ (with an unknown complicated distribution). $\xi$ is further mapped to a sample $X \in \mathcal{X}$ available in $D$. Notice that if the model has sufficient capacity (ample number of hidden layers, as in the case of deep neural networks), the neural network is able to solve the maximization in (1) to obtain $\theta^*$. Let us rewrite our optimization in (1) using $z \sim N(0, I)$ from Theorem (1):

$$\theta^* = \underset{\theta}{\arg\max} \int N(X | f(z; \theta), \sigma^2 * I) N(z | 0, I) dz \quad (3)$$

To solve (3), a distribution function $Q(z | X)$ is defined to decide the importance of an arbitrary configuration $\hat{z} \in Z$ in the generation of a sample $X$. As a consequence, the expected value of $P(X | z)$ with respect to $z$, $\mathrm{E}_{z \sim Q}[P(X | z)]$, can be computed using the Kullback–Leibler (KL) divergence:

$$KL[Q(z) \| p(z | X)] = \mathrm{E}_{z \sim Q}[\log Q(z) - \log P(z | X)] \quad (4)$$

applying the Bayesian rule for $P(z | X)$, (4) can be written as:

$$KL[Q(z) \| p(z|X)] = \mathrm{E}_{z\sim Q}\left[\log Q(z) - \log(\frac{P(X|z)P(z)}{P(X)})\right] \quad (5)$$
$$= \mathrm{E}_{z\sim Q}[\log Q(z) - \log P(X|z) - \log P(z) + \log P(X)]$$

This equality is further written as:
$$\log P(X) - KL[Q(z|X) \| P(z|X)]$$
$$= \mathrm{E}_{z\sim Q}[\log P(X|z) - KL[Q(z|X) \| P(z)]] \quad (6)$$

In order to generate $X$ (that is, create samples $X^* \approx X$), our objective is to maximize $\log P(X)$ while minimizing the KL divergence in the left hand side of (6); hence, we minimize $\mathrm{E}_{z\sim Q}[\log P(X|z) - KL[Q(z|X) \| P(z)]]$ using SGD. Notice that, in the formulation of (6), $Q$ can be viewed as an ANN encoding $X$ into $z$, while $P$ is an ANN decoding $z$ to obtain $X$. To solve the optimization, $Q$ is defined as:
$$Q(z|X) = N(z|\mu(X;\Phi), \Sigma(X;\Phi)) \quad (7)$$
with deterministic functions $\mu$ and $\Sigma$ defined by an ANN with free parameters set $\Phi$ trained by SGD. As $Q$ and $P$ are both $d$ dimensional multivariate Gaussian distributions, the term $KL[Q(z|X) \| P(z)]$ in (6) is computed by:
$$KL[Q(z|X) \| P(z)]$$
$$= KL[N(z|\mu(X;\Phi), \Sigma(X;\Phi)) \| N(0,I)]$$
$$= \frac{1}{2}\left[\log\frac{\det(I)}{\det(\Sigma)} - d + tr(\Sigma) + (0-\mu)^T(0-\mu)\right] \quad (8)$$
$$= \frac{1}{2}\left[-\log(\det(\Sigma)) - d + tr(\Sigma) + \mu^T\mu\right]$$

Therefore, in order to optimize (6), the following optimization problem is solved:
$$\theta^* = \underset{\theta}{\arg\max}\, \mathrm{E}_{X\sim D}\left[\mathrm{E}_{z\sim Q}[\log P(X|z;\Phi)] - KL[Q(z|X;\Phi) \| P(z;\Phi)]\right] \quad (9)$$

Applying the reparametrization technique, (9) can be written as:
$$\theta^* = \underset{\theta}{\arg\max}\, \mathrm{E}_{X\sim D}\left[\begin{array}{l}\mathrm{E}_{\varepsilon\sim N(0,I)}\left[\log P(X|z=\mu(X)+\Sigma^{1/2}(X)*\varepsilon;\Phi)\right] \\ -KL[Q(z|X;\Phi) \| P(z;\Phi)]\end{array}\right] \quad (10)$$

Fig. 4(a) shows the training structure of our generative model based on (7) and (10) to generate $X^* \approx X$. The encoder ANN, $Q$, takes $X$ observed in dataset $D$ and outputs $\mu$ and $\Sigma$ (see (7)). The error of the encoder ANN is $KL[Q(z|X) \| P(z)]$ computed in (8). The gradient of this error function is used by Stochastic Gradient Descent (SGD) method to train this ANN. After computing $\mu$ and $\Sigma$ using $Q$, our latent variable $z = \mu(X;\Phi) + \Sigma^{1/2}(X;\Phi)*\varepsilon$ is obtained using (10). Then, $z$ is fed to the decoder ANN, $P$, to obtain our generated sample $X^* \approx X$. The error function of this ANN is computed by $\|X - X^*\|^2$ to reflect distance between generated sample $X^*$ and its true (observed) value $X$. When $Q$ and $P$ are trained by SGD, in order to generate a new sample $X^* \approx X$, one can simply feed some $z \sim N(0,I)$ to $P$ to obtain $X^*$ as shown in Fig. 4(b).

### B. Convolutional Graph Auto-encoder

In Section III-A, our objective was to learn $P(X)$ in some high dimensional space $\mathcal{X}$ by generating $X^* \approx X$. Here, we aim to learn $P^*(V^*|\pi)$, i.e. PDF of $V^*$ in $G$ given $\pi$. We present our CGAE shown in Fig. 5 as the first generative model that captures nodal distribution $P^*(V^*(t')|\pi)$ in a graph $G$. Given historical GHI $\pi$, our objective is to generate $\rho$ samples $\hat{V} \approx V^*$ to estimate $P^*(V^*|\pi)$.

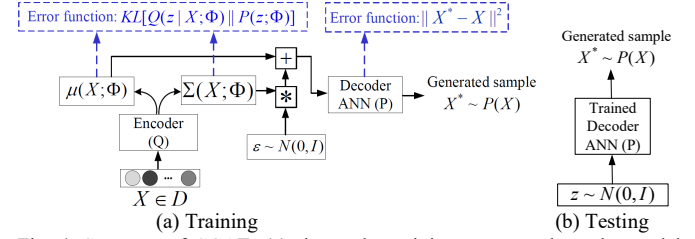

(a) Training (b) Testing

Fig. 4. Structure of CGAE. (a) shows the training process where the model generates $X^* \approx X$. (b) shows the testing process where the trained decoder generates as many samples $X^* \sim P(X)$ as required simply by feeding a random $z \sim N(0,I)$ to the decoder ANN. The decoder captures PDF $P(X)$.

Let us mathematically formalize how CGAE generates $\hat{V}$ as an estimation for $V^*$:
$$\hat{V} = \mu(\pi,z) + \varepsilon \quad s.t.\quad z \sim N(0,1),\, \varepsilon \sim N(0,1) \quad (11)$$
both $z$ and $\varepsilon$ are white Gaussian noises. $\mu$ is implemented by an ANN as in Section III-A. Assuming $z \sim Q$ using PDF $Q(z)$, Bayes rule [20] is applied to compute $\mathrm{E}_{z\sim Q}[\log P(V^*(t')|z,\pi)]$:
$$\mathrm{E}_{z\sim Q}[\log P(V^*(t')|z,\pi)] = \mathrm{E}_{z\sim Q}[\log P(z|V^*(t'),\pi) - \log P(z|\pi) + \log P(V^*(t')|\pi)] \quad (12)$$

(12) is rewritten as:
$$\log P(V^*(t')|\pi) - \mathrm{E}_{z\sim Q}[\log Q(z) - \log P(z|\pi,V^*(t'))] = \mathrm{E}_{z\sim Q}[\log P(V^*(t')|z,\pi) + \log P(z|\pi) - \log Q(z)] \quad (13)$$

Now, following (7), we have $Q = N(\mu'(\pi,V^*(t')), \sigma'(\pi,V^*(t')))$ where $\mu'$ and $\sigma'$ are ANNs trained alongside $\mu$. Let us denote $Q$ by $Q(z|\pi,V^*)$, (13) is written as:
$$\log P(V^*|\pi) - KL[Q(z|\pi,V^*) \| P(z|\pi,V^*)] = \mathrm{E}_{z\sim Q}[\log P(V^*|z,\pi)] - KL[Q(z|\pi,V^*) \| P(z|\pi)] \quad (14)$$

Considering (14), our objective is to increase $E_1 = \log P(V^*|z,\pi)$ and $E_2 = -KL[Q(z|\pi,V^*) \| P(z|\pi)]$. CGAE is trained by SGD to maximize $E_T = E_1 + E_2$. This leads to maximizing the likelihood of $V^*$ while training $Q$ to accurately estimate $P(z|\pi,V^*)$. Note that, similar to our optimization in Section III-A, we have $P(z|\pi) = N(0,1)$. Our latent vector is $z = \mu'(\pi,V^*(t')) + \alpha \circ \sigma'(\pi,V^*(t'))$ where $\alpha \sim N(0,1)$ and $\circ$ is the element-wise product operation. $E_T$ is differentiable with respect to the whole parameters of CGAE (including the parameters in ANNs corresponding to $\mu$, $\mu'$ and $\sigma'$); hence, the whole CGAE model can be easily tuned by SGD to maximize $E_T$. In Section III-C, the neural architecture corresponding to our CGAE is defined based on ANNs.

### C. CGAE Architecture

CGAE consists of three ANNs; 1- Graph Feature Extraction ANN, which gives us a compact representation of $\pi$ stored in $G$, denoted by $R(G)$, 2- Encoder ANN, $Q$, that implements $\mu'$ and $\sigma'$ to capture $Q(z|\pi,V^*)$, and 3- Decoder ANN, $P$, that implements $\mu(\pi,z)$ in (11), to produce samples $\hat{V}$ drawn from the true future GHI distribution $P^*(V^*(t')|\pi)$.

*1) Graph Feature Extraction ANN (Computing R(G))*

At each training step $t$, the spectral graph convolutions of $G$, which stores $\pi = <T(v_1,t),T(v_2,t),...,T(v_n,t)>$ inside its nodes, is computed by $\psi_\theta * \pi = U\psi_\theta U^T \pi$. Here, $U$ is the eigenvector matrix of the normalized Laplacian $L = U\Omega U^T$ and $\theta \in \mathbb{R}^n$ is the parameter vector for the convolutional filter $\psi_\theta = diag(\theta)$ in the Fourier domain. Notice that the Fourier transformation of $\pi$ is computed by $U^T \pi$. $\psi_\theta$ is defined as a function of $L$'s eigenvalues; hence, our filter is denoted by $\psi_\theta(\Omega)$. Estimating $\psi_\theta(\Omega)$ by Chebyshev Polynomials [21] $P_j$, we have $\psi_\omega \approx \sum_{j=0}^{J} \omega_j P_j(\frac{2}{\gamma_{max}}\Omega - I)$ where $\gamma_{max}$ is maximum eigenvalue of $L$, and $\omega_j$ is the $j$-th Chebyshev coefficient. Therefore, the spectral graph convolution function on $G$ is:

$$\psi_\omega * \pi \approx \sum_{j=0}^{J} \omega_j P_j(\frac{2}{\gamma_{max}}\Omega - I)\pi \quad (15)$$

The convolution in (15) is further simplified by $\delta = \omega_0 = -\omega_1$ which decreases parameters' size while $\gamma_{max} = 2$ for $J = 1$; As a result, (15) can be computed by:

$$\psi_\omega * \pi \approx \omega_0 P_0(L-I)\pi + \omega_1 P_1(L-I)\pi = \delta(I + D^{-\frac{1}{2}}AD^{-\frac{1}{2}})\pi \quad (16)$$

Based on the convolution (16), a graph feature extraction neural network (GFENN) with $L_G$ hidden layers is defined to extract spatio-temporal features from GHI observations at all nodes/sites of $G$. Here, the output of each layer $1 \leq k \leq L_G$ is:

$$O^k = \text{ReLU}(MO^{k-1}W^k) \quad s.t. \quad M = \tilde{D}^{-\frac{1}{2}}(A+I)\tilde{D}^{-\frac{1}{2}} \quad (17)$$

where $\tilde{D}_{ii} = \sum_j (A+I)_{ij}$. The input of GFENN is $O^0 = \pi$ while the output is $G$'s spatio-temporal representation $R(G) = O^{L_G}$.

*2) Encoder (Q) and Decoder (P)*

Since GFENN captures spatiotemporal features of $\pi$, and stores them in $R(G)$, one can view CGAE as a model estimating $P^*(V^*|R(G))$ instead of $P^*(V^*|\pi)$. In Section III-A, (7) showed that $Q$ can be viewed as an ANN encoding input tensor $X$ into the latent vector $z$ while $P$ is a decoding ANN that maps $z$ to $X$. As depicted in Fig. 5, Here, the input to the encoder $Q$ is $X = R(G)$. Our encoder $Q$ is defined by a deep ANN with $L_Q$ hidden layers and ReLU activations for each hidden layer, trained to encode $V^*$ into latent vector $z \in Z$, such that the resulting $z$ can be decoded back to $V^*$. As discussed in (14) and also shown in Fig. 5, the error function for the encoder $Q$ is defined by:

$$Err_Q = KL[Q(z|\pi,V^*) \| N(0,1)] = KL[Q(z|R(G),V^*) \| N(0,1)] \quad (18)$$

Similar to $Q$, our decoder, $P$, is implemented by a deep ANN with $L_P$ hidden layers using ReLU activations to take the latent vector $z$ learned by $Q$, as well as the graph representation $R(G)$, and decode them to generate an approximation of $V^*$, denoted by $\hat{V}$. To make the generated sample $\hat{V}(t')$, as close as possible to the real future value $V^*(t')$ we minimize the following reconstruction error for $P$:

$$Err_P = \| V^*(t') - \hat{V}(t') \|^2 \quad (19)$$

Therefore, the total error optimized by the stochastic gradient descent method is $E = Err_Q + Err_P$.

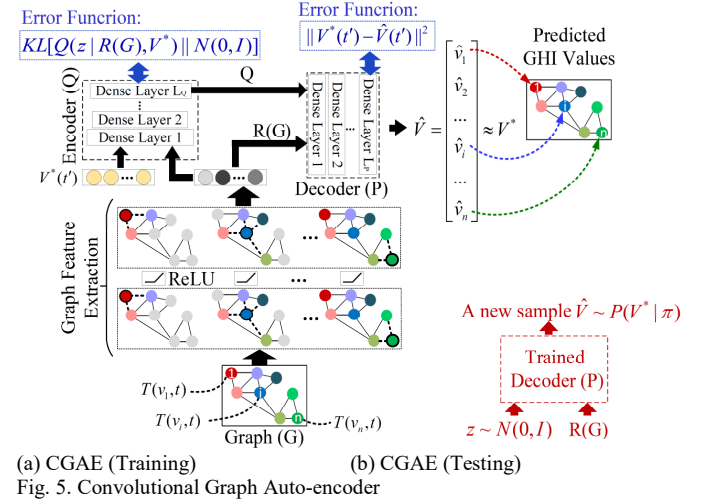

(a) CGAE (Training)  (b) CGAE (Testing)
Fig. 5. Convolutional Graph Auto-encoder

*D. Estimation of $P(V^*|\pi)$*

As shown in Fig. 5(b), during test time, $R(G)$ and $z \sim N(0,I)$ are fed to the decoder ANN and the estimation $\hat{V}(t')$ is obtained. No encoding is needed; hence, generating estimations $\hat{V}(t') \approx V^*(t')$ is dramatically fast. All we need to do to generate a new sample $\hat{V}(t')$, is to sample a new $z \sim N(0,I)$ and run feed-forward algorithm on the GFENN (to obtain $R(G)$) and the decoder ANN (to obtain the desired result, i.e. $\hat{V}(t')$). Following this approach, we generate $\rho$ number of samples $\hat{V} \sim P(V^*|\pi)$ to estimates $P(V^*|\pi)$ using the decoder. As a result, our decoder $P$ generates the PDF of future GHI mapping $N(0,I)$ to $P(V^*|\pi)$.

IV. NUMERICAL RESULTS

CGAE is compared with recent benchmarks for short-term irradiance/PV probabilistic forecasting: Persistence Ensemble (PEn) [22], Quantile Regression (QR) [19], Kernel Density Estimation (KDE) [18], and Extreme Learning Machines (ELM) [16, 17]. The advantages of spatio-temporal feature learning for the underlying problem is shown. Since no generative model was presented in the literature, the experiments motivate further research on generative modeling for renewable resources prediction.

## A. Experimental Settings

As explained in Section II, the NSRD dataset is applied to train/test our model. The 1998-2015 data is used to train CGAE while 2016 data is applied to evaluate the prediction performance. In this study, CGAE is trained/tested to forecast GHI time series from 30 min (horizon length $k=1$) up to 6 hours ahead ($k=12$). Stochastic Gradient Descent with learning rate $\eta = 5*10^{-4}$ is employed to train our CGAE (including GFENN, encoder ANN, and decoder ANN) by decreasing the error $Err_Q + Err_P$. In this study, the number of generated samples is $\rho = 10^4$, and the number of GFENN layers is set to $L_G = 2$ while $L_P = 4$ and $L_Q = 3$. The feature selection hyper parameter is $\tau = 0.45$. The model is implemented on a computer system with Intel Core-i7 4.1GHz CPU and NVIDIA GeForce GTX 1080-Ti GPU.

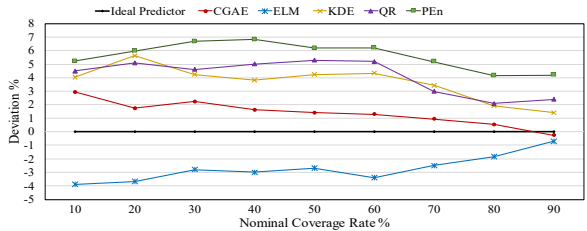

Fig. 6. Reliability measurements averaged over all GHI nodes/sites

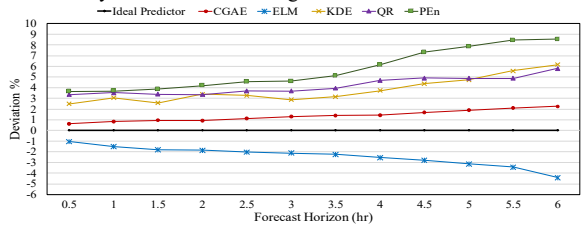

Fig. 7. Average reliability with different look-ahead times

## B. Performance Comparison (Quantitative Results)

The prediction quantiles of our model are compared with other methodologies in terms of reliability, sharpness and Continuous Ranked Probability Score (CRPS):

*1) Reliability*: This criterion shows how closely the prediction probabilities correspond to the observed (real) frequencies of the GHI data. Here, the bias $R^{1-2\alpha}$ is computed by:

$$R^{1-2\alpha} = \left(\frac{N^{1-2\alpha}}{N} - (1-2\alpha)\right) \times 100\% \quad (20)$$

where $N$ is the number of test examples, $N^{1-2\alpha}$ is the number of observations covered by the nominal coverage rate $(1-2\alpha) \times 100\%$. The closer the nominal coverage of prediction intervals is to the observed (actual) coverage rate, the higher the reliability is; hence, small $R^{1-2\alpha}$ shows better accuracy. In fact, $R^{1-2\alpha}=0$ corresponds to the perfect (ideal) reliability. Fig. 6 depicts the reliability measurements averaged over all GHI nodes/sites with various nominal coverage rates ranging from 10% to 90%. In Fig. 6, CGAE has the closest curve to the optimal curve, with an average absolute deviation of 1.45%, while ELM is the second best approach with 2.71% deviation. QR and KDE have relatively close performance in 10%-30% and 70%-80% rates; however, KDE shows better performance in other ranges. Both KDE and QR have noticeable improvement compared to PEn in 30%-90% range. The ANN-based models, CGAE and ELM, improve the average absolute reliability of KDE by 0.61% and 0.26%, respectively. This shows the advantage of applying nonlinear standard function approximation for the prediction problems. Fig. 7 shows the average reliability with different look-ahead times. As shown in this plot, the slope of the deviation curve for all benchmarks start to increase significantly from the 3.5-hr horizon, while CGAE has a much smaller slope. As the time horizon expands, the improvement of CGAE becomes more significant. PEn has the worst performance, especially in longer horizons, compared to other methodologies. This is due to its low generalization capacity resulted from its smoothness assumption of the target function, which undermines its efficiency in practice. Both ANN-based approaches, CGAE and ELM have less than 5% reliability measure for all time horizons while PEn, KDE, and QR exceed this limit. CGAE yields 0.39% and 0.49% better reliability in 3-hr and 6-hr forecasts compared to ELM. This shows the superiority of generative modeling over discriminative modeling introduced in previous ANN methods. The small deviation of CGAE is resulted by a good unbiased prediction, while other models are more biased which decreases their efficiency in practical applications.

*2) Sharpness:* Sharpness is a complimentary metric to the reliability, which evaluates the concentration of the prediction distribution. The criterion shows how informative a forecast is by narrowing down the predicted GHI values. Sharpness should be analyzed with respect to reliability, as high sharpness does not necessarily show better prediction when the model has low reliability (high deviation in Fig. 6 and 7). Sharpness is investigated using two performance metrics:

- **Prediction Interval Average Width (PIAW):**

This metric, $PIAW^\alpha$, evaluates sharpness for the nominal coverage rate $(1-2\alpha) \times 100\%$ by:

$$PIAW_\alpha = \frac{1}{N}\sum_{n=1}^{N}|q^\alpha(n) - q^{1-\alpha}(n)| \quad (21)$$

where $q^\alpha(n)$ and $q^{1-\alpha}(n)$ represent the $\alpha$ and $1-\alpha$ prediction quantiles for the $n$-th test sample. Fig. 8 shows the average sharpness of 10%-90% nominal coverage rates normalized by maximum observed GHI. As shown in this diagram, PEn has the sharpest intervals in all nominal coverage rates; however, as shown by Fig. 6 and 7, it has poor reliability compared to other benchmarks especially when the horizon is expanded. ELM provides overly narrow quantiles leading to higher sharpness compared to CGAE. However, its sharpness does not contribute to forecast accuracy/reliability. Such high sharpness might work in the case of clear sky when no significant uncertainty is present and GHI is predictable with high accuracy; however, in other cases (e.g. when GHI is varying during a rainy day), it would lead to poor performance as ELM would neglect the risk of uncertainties in GHI. CGAE provides medium sharpness which is not too high to lead to erroneously narrow quantiles (as in the case of PEn and ELM), and not too low to lose information about future GHI (as in the case of KDE and QR).

- **PDF Entropy:**

Sharpness of a forecast can be estimated using the entropy of the prediction PDF. Sharper forecasts have smaller PDF entropies. Fig. 9 shows the histogram of the entropies for all benchmarks in 6-hr ahead forecasting task. As shown here, the majority of forecasts for PEn and ELM correspond to low values. The mean entropy of PEn and ELM are 2.77 and 3.69, respectively. The low entropy of PEn is due to the consecutive clear days where the variance of the prediction PDF is small. Such small entropies/variances result in overconfident predictions caused by lack of knowledge about the future GHI uncertainty. The overly narrow prediction quantiles in ELM lead to low PDF entropies which degrades accuracy as the uncertainties in the future GHI is disregarded by predictions less reliable than CGAE (see Fig. 6 and Fig.7). CGAE has moderate sharpness and medium entropy values with mean 5.15. KDE has high entropies with mean 6.77, and a small variance of 0.22 that result in high uncertainty boundaries for the future GHI and less informative forecasts compared to CGAE and ELM. In contrast to ELM and KDE, our CGAE model has entropies not too low (as in ELM) to disregard GHI uncertainties and not too high (as in KDE) to provide underconfident predictions.

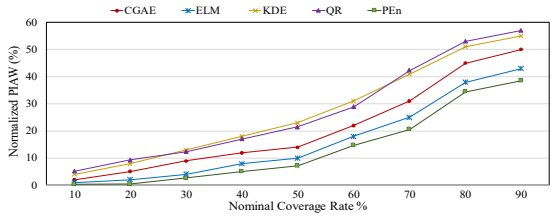
Fig. 8. Sharpness evaluation using normalized PIAW

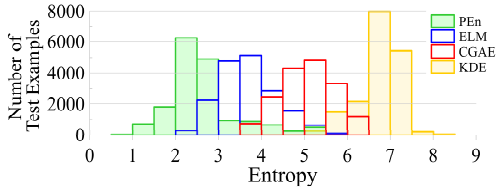
Fig. 9. Entropy diagram of various benchmarks for 6-hr ahead forecasts

*3) Continuous Ranked Probability Score:*

CPRS is a metric evaluating the entire prediction distribution reflecting the deviations between the CDF of the predicted and observed data. One can view CRPS as a metric combining reliability and sharpness to provide a comprehensive performance evaluation. CRPS is computed by:

$$CRPS(F,v) = \int_{-\infty}^{\infty} (F(x) - U(x-v))^2 dx \text{ with } U(x) = \begin{cases} 1 & x \geq 0 \\ 0 & x < 0 \end{cases} \quad (22)$$

with the prediction CDF $F$ and the Heaviside function $U$. The average CRPS for 30-min up to 6-hr ahead GHI forecast is depicted in Fig. 10. The smaller CRPS a model yields, the better accuracy it provides. As shown in this plot, the ANN-based methodologies, ELM and CAGE, outperform PEn, QR, and KDE. ELM achieves 12% and 16% better CRPS on average over all time horizons compared to KDE and QR, respectively. KDE has slightly better performance in comparison with QR for 30-min up to 2.5-hr ahead predictions. The better accuracy of KDE becomes more noticeable in the horizon range of 3 hr up to 4.5 hr. Similar superiority is also reflected by the better reliability curve of KDE compared to QR in Fig. 7. Among all benchmarks, PEn has the worst performance. PEn has 14% and 9% more CRPS on average, compared to KDE and QR, respectively. As the forecast horizon length grows, the CRPS of PEn increases by larger amounts compared to other benchmarks. This is due to low generalization capability and erroneously high sharpness (low entropy as shown in Fig. 9) which results in unreliable predictions, especially when the weather condition changes from sunny to cloudy since this approach suffers from the naïve smoothness assumption. As

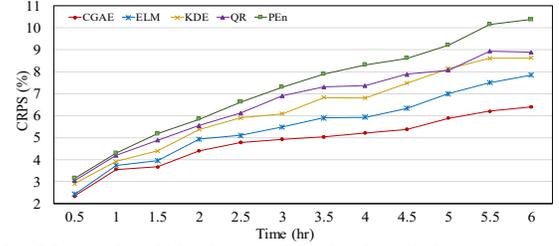
Fig. 10. CRPS results of 30-min up to 6-hr ahead predictions

depicted in Fig. 10, CGAE shows best performance because of its high reliability (shown by Fig. 6-7) and appropriate sharpness (i.e., moderate PIAW and entropy in Fig. 8-9). CGAE outperforms ELM by 4.8% CRPS for hourly predictions which is increased significantly for time horizons of length more than 3 hours, and reaches the 18% CRPS improvement for 6-hr predictions.

*C. Qualitative Results*

The probabilistic prediction of CGAE is investigated to show the capability of our model under different weather conditions. Fig. 11(a) shows the GHI values of eight days, from June 25th to July 2nd in 2016, for a site near the Michigan Lake. As shown in this plot, the selected days contain various weather conditions including sunny, partly cloudy, and overcast, in a short period of time. June 25th and 26th are both sunny with high GHI, while the subsequent day, June 27th, is mostly cloudy with many variations. The next day, June 28th is sunny with high GHI while June 29th is overcast with very small irradiance. June 30th and July 1st are sunny, and the last day, July 2nd is a combination of partly cloudy and sunny. This test case evaluates the performance of CGAE when the weather changes dramatically from one day to the other, and within each day. As shown in Fig. 11 (b)-(e), the prediction intervals of CGAE with 50% and 90% confidence rates follow the actual GHI values with high accuracy resulting in good reliability. In Fig. 11(b), as the weather changes from sunny to partly cloudy around 9:00, the confidence boundaries expand showing the increase in the prediction uncertainty. In Fig. 11(c), June 28th has a very smooth GHI curve measured on a clear sunny day, hence, the model's uncertainty is very small. In Fig. 11(d)-(e) the weather has significant changes during overcast in June 29th and partly cloudy and sunny conditions in July 2nd. As seen in these two figures, although the uncertainty is increased in such conditions, the model still follows the observed GHI with high reliability. On July 2nd, at 12:30, the GHI jumps drastically from 12% of maximum GHI, $GHI_{MAX}$, to 86%. Fig. 11(f) shows the histogram of the predicted GHI for this observation. As shown in this figure, CGAE could capture this jump more reliably having heavier probability density around 85%-90% $GHI_{MAX}$. However, ELM and KDE assign high probability to smaller values as these models are more affected by previous small measurements. Moreover, KDE does not provide enough sharpness for this example, hence, its prediction cannot be informative. Having much higher generalization capability and being able to leverage spatio-temporal information from GHI observations, our CGAE can capture uncertainties in the solar data with higher accuracy and appropriate sharpness.

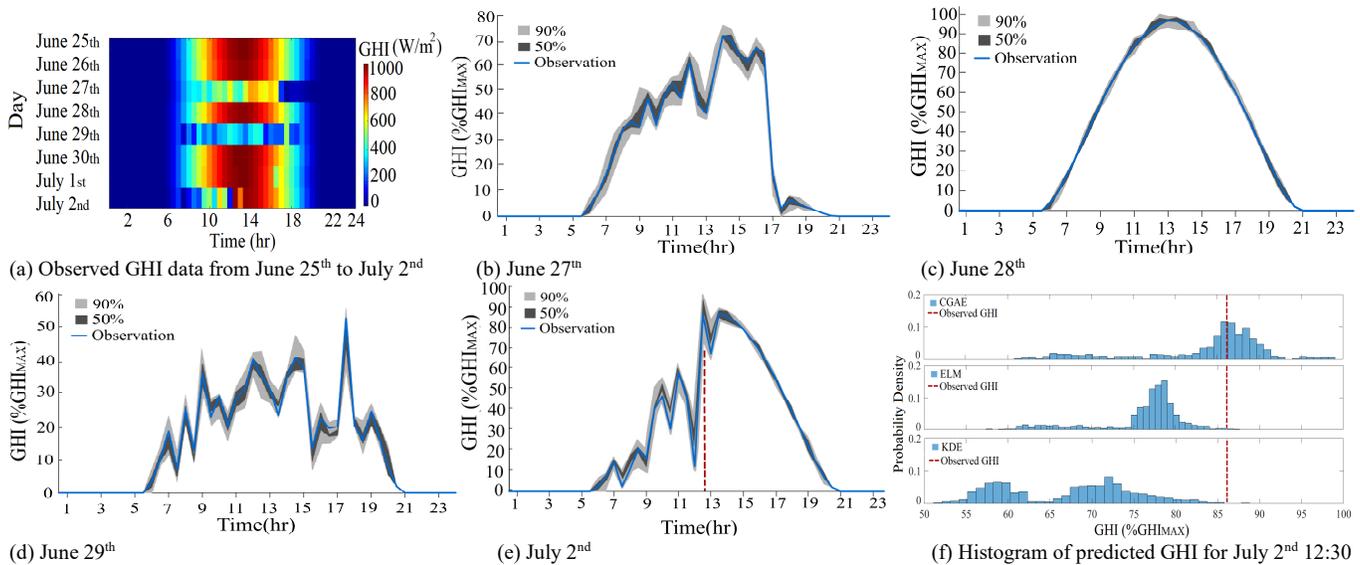

Fig. 11. Predicted densities forecasted by CGAE in four days between June 25th and July 2nd with various weather conditions

## V. CONCLUSIONS

A novel deep generative model, Convolutional Graph Auto-encoder, is presented for a new problem, nodal distribution learning in graphs. The model captures deep convolutional features from an arbitrary graph-structured data, to learn the corresponding probability densities of nodes. Here, the problem of spatio-temporal solar irradiance forecasting is presented as a graph distribution learning problem where each node of the graph represents a solar irradiance measurement site, while each edge represents the distance between the sites. Using graph spectral convolutions, the spatial features of the solar data are extracted, that are further used by an encoding and decoding ANN to capture the distribution of future solar irradiance. Our deep learning model is used to provide probabilistic forecasts for the National Solar Radiation Database. Simulation results show better reliability, sharpness and Continuous Ranked Probability Score compared to recent baselines in the literature.